\documentclass[preprint,12pt]{elsarticle}

\makeatletter
\def\ps@pprintTitle{%
 \let\@oddhead\@empty
 \let\@evenhead\@empty
 \def\@oddfoot{\centerline{\thepage}}%
 \let\@evenfoot\@oddfoot}
\makeatother

\usepackage{amssymb}
\usepackage{amsmath}

\usepackage{url} 
\usepackage{subcaption}  
\usepackage{hyperref}
\usepackage{geometry}
\usepackage{booktabs}

\journal{Elsevier}

\begin{document}

\begin{frontmatter}



\title{Automating Traffic Monitoring with SHM Sensor Networks via Vision-Supervised Deep Learning}


\author[ibk]{Hanshuo Wu}
\author[ibk,sec]{Xudong Jian}
\author[irmos]{Christos Lataniotis}
\author[ibk,irmos]{Cyprien Hoelzl}
\author[ibk,sec]{Eleni Chatzi\corref{cor1}}
\author[ibk,irmos]{Yves Reuland}

\cortext[cor1]{Corresponding author. Email-address: chatzi@ibk.baug.ethz.ch}

\affiliation[ibk]{organization={Institute of Structural Engineering (IBK), ETH Zürich},
            city={Zürich},
            country={Switzerland}}
\affiliation[irmos]{organization={irmos technologies AG},
            city={Zürich},
            country={Switzerland}}
\affiliation[sec]{organization={Future Resilient Systems, Singapore-ETH Centre},
            city={Singapore},
            country={Singapore}}

\begin{abstract}
Bridges, as critical components of civil infrastructure, are increasingly affected by deterioration, making reliable traffic monitoring essential for assessing their remaining service life. Among operational loads, traffic load plays a pivotal role, and recent advances in deep learning—particularly in computer vision (CV)—have enabled progress toward continuous, automated monitoring. However, CV-based approaches suffer from limitations, including privacy concerns and sensitivity to lighting conditions, while traditional non-vision-based methods often lack flexibility in deployment and validation. To bridge this gap, we propose a fully automated deep-learning pipeline for continuous traffic monitoring using structural health monitoring (SHM) sensor networks. Our approach integrates CV-assisted high-resolution dataset generation with supervised training and inference, leveraging graph neural networks (GNNs) to capture the spatial structure and interdependence of sensor data. By transferring knowledge from CV outputs to SHM sensors, the proposed framework enables sensor networks to achieve comparable accuracy of vision-based systems, with minimal human intervention. Applied to accelerometer and strain gauge data in a real-world case study, the model achieves state-of-the-art performance, with classification accuracies of 99\% for light vehicles and 94\% for heavy vehicles.

\end{abstract}

\begin{graphicalabstract}
    \centering
    \includegraphics[width = 0.7\textwidth]{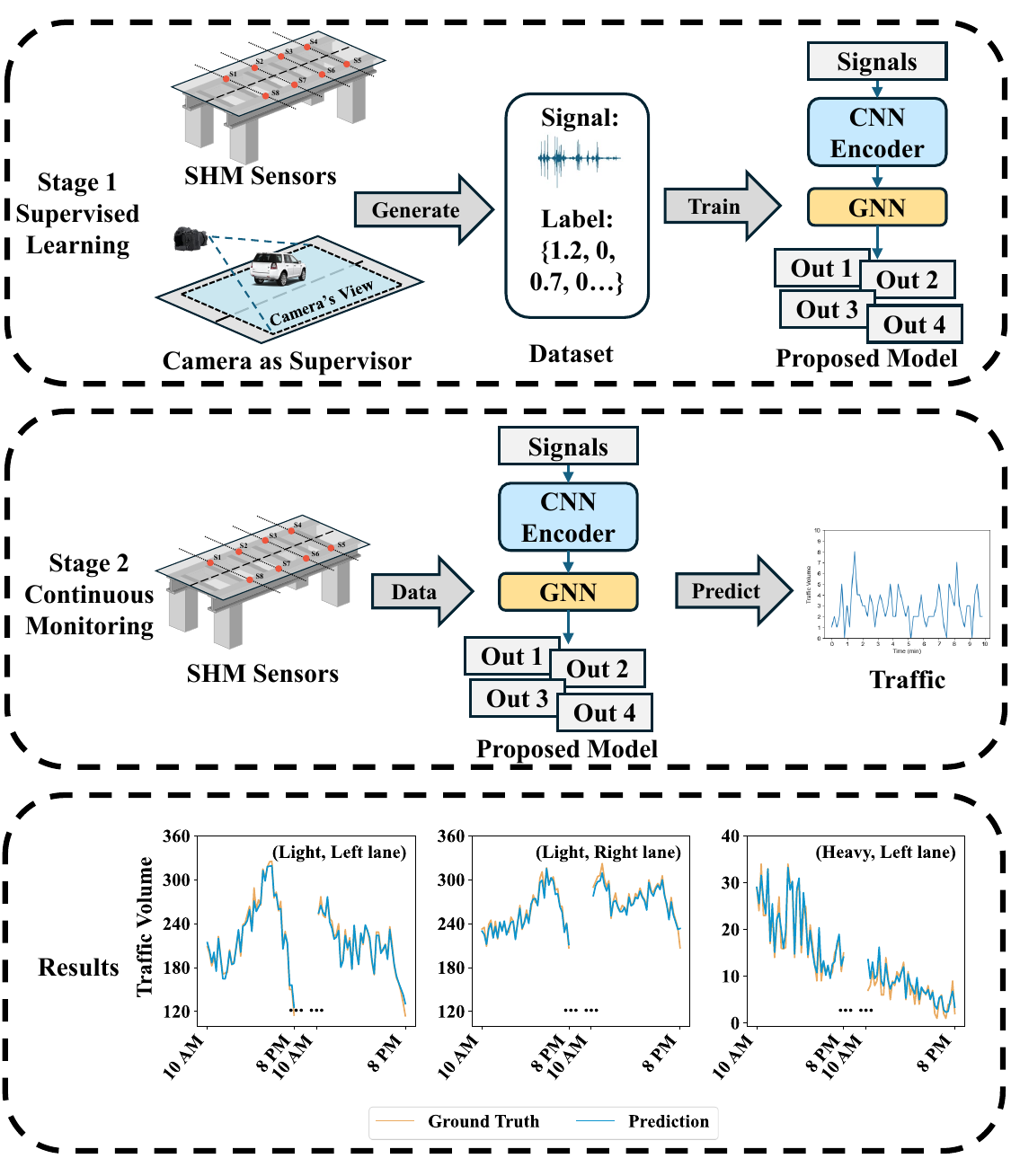}
\end{graphicalabstract}

\begin{highlights}
\item We propose a fully automated traffic monitoring pipeline using SHM sensor networks, trained with data labeled by computer vision models, requiring no road-level installation or manual signal annotation.

\item Our model, with a graph neural network decoder, achieves state-of-the-art performance: 99\% accuracy for light vehicles and 94\% for heavy vehicles.

\item We validate our method on a two-lane highway bridge with one-week synchronized video and SHM data, providing vehicle type, speed, lane, and structural response.
\end{highlights}


\end{frontmatter}



\section{Introduction}\label{sec1}

Bridges are vital assets within civil infrastructure systems, playing a central role in supporting economic activity and connecting communities \citep{jimenez_rios_bridge_2023}. However, they are increasingly threatened by progressive deterioration. Since the beginning of the 21st century, over 120 bridges collapsed worldwide, causing significant economic losses and serious societal impacts \citep{wang_installation_2022}. This deterioration arises from multiple contributing factors, including material aging, exposure to harsh environmental conditions, and rising operational demands. Among these, traffic loads represent the predominant form of operational loading and are a key contributor to the accumulation of fatigue damage over time \citep{yang_fatigue_2022}. With the rapid increase in traffic volume in recent years \citep{nellore_survey_2016, yang_traffic_2019}, the role of traffic loading has become even more critical in evaluating bridge performance, highlighting the importance of continuous traffic load monitoring as part of effective bridge management strategies.

Traffic volume monitoring, or traffic monitoring, is a transportation network-level problem, and comprises several tasks, including detection, counting, classification, and weight estimation, each varying in difficulty and complexity \citep{premaratne_comprehensive_2023}. This study will focus on vehicle counting and classification in traffic monitoring problems.
From the perspective of this work, traffic monitoring methods can be classified into two solutions: vision-based and non-vision-based, respectively. Here, non-vision-based solutions, or contact-based solutions, refer to solutions using sensors-typically deployed on the structure, such as accelerometers, strain gauges, and magnetometers, but do not include cameras or infrared cameras.

Vision-based solutions typically rely on advanced computer vision (CV) models, which process imagery captured by on-site surveillance cameras—either temporarily or permanently installed. These technologies offer several notable advantages, such as high spatial resolution and the ability to extract detailed traffic information, which have contributed to their growing popularity among researchers and practitioners in recent years\citep{chowdhury_real_2018,jian_integrating_2022,basheer_ahmed_real-time_2023}. CV systems are relatively easy to deploy, require only a low-cost camera as on-site equipment, and typically deliver accurate and robust results. However, they also present several critical limitations. These include potential data security and privacy concerns, reduced performance under low-light or adverse weather conditions, and the inability to infer key attributes such as vehicle weight (e.g., whether a truck is loaded) from visual data alone.


Non-vision-based, or sensor-based, traffic monitoring solutions utilize data from various sensors such as accelerometers \citep{burrello_traffic_2022} and strain gauges \citep{paul_application_2023}, which are typically installed on bridges for structural health monitoring (SHM). Although not originally intended for traffic monitoring, SHM sensors capture key responses to vehicular activity—for instance, passing vehicles generate measurable vibrations and strain. Several studies have shown that dynamic axle forces can be estimated from acceleration data, enabling bridge weigh-in-motion (B-WIM) applications \citep{mustafa_iterative_2021, wang_acceleration_2020}.

Unlike visual data, which can often be interpreted with minimal preprocessing, sensor data are less intuitive and require more complex signal processing. Techniques range from unsupervised approaches (e.g., feature engineering, peak detection) to supervised machine learning methods. Prior research has demonstrated the effectiveness of both strategies \citep{burrello_traffic_2022, burrello_enhancing_2020}, though unsupervised methods often suffer from limited accuracy. While supervised learning improves performance, it remains challenged by complex conditions such as multi-lane traffic scenarios, where load effects are spatially coupled. For example, a heavy vehicle in one lane may induce significant responses in adjacent lanes. To date, no study has investigated the use of deep learning for traffic monitoring using SHM sensors under such multi-lane conditions, leaving a critical gap in the literature.

Given their respective strengths, vision-based and non-vision-based methods are each suited to different traffic monitoring scenarios and requirements. Non-vision-based solutions, in particular, address several key limitations of vision-based approaches: they do not raise privacy concerns, are unaffected by lighting or weather conditions, and—most importantly—can capture structural responses indicative of vehicle weight, especially when using accelerometers and strain gauges. These capabilities make them particularly valuable for applications requiring load estimation.
However, sensor-based methods also present notable challenges. They rely on indirect inference of traffic loads, making signal interpretation more complex, especially in scenarios involving overlapping signals from multiple vehicles on long or multi-lane bridges. Their deployment typically involves higher costs, though these can be offset when traffic monitoring is integrated with broader SHM objectives. In cases where accurate load estimation is needed, additional system calibration through load testing is required, which can be resource- and cost-intensive.
Despite their growing appeal, non-vision-based solutions remain constrained in terms of scalability due to their complexity, sensor costs, and lower reliability relative to visual systems. A comparative summary of the two approaches is provided in \autoref{tab:method_comparison}.

\begin{table}[htp!]
\centering
\begin{tabular}{l|c|c}
\hline
 Benefits & Vision-based methods & Sensor-based methods \\ \hline
Privacy protection                & – & + \\ 
Insensitivity to lighting            & – & + \\ 
Weight information              & – & + \\ 
Low cost equipment                  & + & + (with SHM) \\ 
Easy deployment            & + & – \\ 
Accuracy            & + & – \\ 
\hline
\end{tabular}
\caption{Comparison of advantages (+) and disadvantages (–) of vision‑based versus non‑vision‑based traffic monitoring methods.}
\label{tab:method_comparison}
\end{table}

In this study, we aim to harness the complementary strengths of vision-based and sensor-based approaches by utilizing SHM sensors for traffic monitoring tasks. We demonstrate that SHM sensor networks, pre-trained on data annotated by CV models, can achieve accuracy comparable to that of the CV models themselves, while offering advantages in terms of reduced system complexity and minimal human intervention. We formulate traffic monitoring as a multi-task regression problem, where the input consists of typical SHM signals—such as acceleration and strain—and the output includes both the number of vehicles by type and their corresponding lane assignments. The main contributions of this study are:

\begin{itemize}
\item We propose a fully automated deep learning pipeline for traffic monitoring using SHM sensor networks, encompassing CV-assisted dataset generation, supervised training, and inference. Our results demonstrate that sensor networks can effectively learn from CV outputs and achieve accuracy comparable to vision-based methods using both basic and deep regression models. Unlike vision-based systems, our approach avoids privacy and data security concerns, operates reliably under varying lighting conditions, and requires no equipment on the road surface. Furthermore, it minimizes human intervention, as no manual signal annotation is needed, significantly reducing the cost and effort required for deployment on additional bridges.

\item We show that among the models tested, a graph neural network (GNN)-based decoder yields the best performance in interpreting sensor network data. Comparative evaluations against several machine learning approaches and existing studies confirm that our GNN-based model achieves state-of-the-art results for sensor-driven traffic monitoring.

\item As a proof of concept, we validate our framework using a case study involving a two-lane highway bridge in Switzerland. Over a one-week period, we collected high-resolution data including vehicle events, trajectories, types, speeds, lane positions, and synchronized acceleration and strain signals.
\end{itemize}


\section{Related Works}\label{sec2}
\subsection{Vision-based traffic monitoring}
Vision-based traffic monitoring is commonly framed as a standard object detection problem. Typically, this involves the installation of a surveillance camera on-site and the application of an object detection algorithm to the video frames captured. Among the most widely adopted models are the YOLO (You Only Look Once) series, which have undergone rapid development, with the latest version being YOLOv12 \citep{tian_yolov12_2025}. YOLO models are designed for real-time object detection and are capable of identifying a wide range of objects—including vehicles, pedestrians, and animals—across diverse scenarios. Their favorable trade-off between inference speed and detection accuracy has made them particularly popular in civil engineering applications, such as crack detection, construction site monitoring, and traffic counting. In addition to the YOLO family, several other high-performance object detection models have recently emerged, including RT-DETR \citep{zhao_detrs_2024} and RTMDet \citep{lyu_rtmdet_2022}, further advancing the capabilities of vision-based monitoring systems.


In traffic monitoring applications, object tracking algorithms are commonly applied to sequences of video frames following vehicle detection, enabling vehicle trajectory reconstruction and speed estimation. Among multi-object tracking methods, the ByteTrack and SORT (Simple Online and Realtime Tracking) families are widely adopted due to their efficiency and simplicity \citep{zhang_bytetrack_2022, bewley_simple_2016, aharon_bot-sort_2022, cao_observation-centric_2023}. These trackers typically employ Kalman filters to predict object motion trajectories, subsequently associating the predicted bounding boxes with those obtained from object detection to maintain continuous tracking. While some tracking algorithms also incorporate visual appearance features to improve association accuracy, this often comes at the expense of computational efficiency \citep{wojke_simple_2017}.
With object tracking algorithms, the speed of vehicles can be obtained, as long as the Euclidean dimension in the frame is known. The speed can be calculated by measuring the time required for a vehicle to cross a fixed distance within the camera view \citep{luo_computer_2023}. The axle count of vehicles can also be determined from visual data by detecting and counting the number of wheel objects within the bounding box of a vehicle \citep{zhu_fine-grained_2022}.

In order to record the trajectory of a vehicle on a bridge, camera calibration is needed to build the mapping between pixels and world coordinates. \citet{zhu_fine-grained_2022} use a large chessboard placed on the road for camera calibration. However, such setups are often impractical or inconvenient in real-world deployments. To address this, a simplified approach that assumes the bridge deck forms a flat plane is widely used \citep{dong_large_2023, jian_bridge_2024, yin_traffic_2023}. Under this assumption, the mapping can be constructed using at least four known control points visible in the camera's field of view. This effectively reduces the 2D-to-3D projection to a planar homography, allowing each detected object in the video frames to be projected into world coordinates and enabling accurate trajectory estimation.

Beyond conventional object tracking, some studies have explored the fusion of visual data with vehicle weight information obtained from B-WIM systems. This approach typically requires tracking individual vehicles across multiple camera views distributed along a long-span bridge, allowing the weight data from the weigh-in-motion (WIM) system to be associated with the corresponding vehicle \citep{dong_large_2023, zhou_hybrid_2024}. As a result, a more comprehensive spatiotemporal representation of bridge traffic can be achieved. \citet{yang_addressing_2024} achieved a more accurate real-time detection solution for vehicle wheel weight by fusing wheel location identification with CV techniques and the WIM system. Other studies have attempted to estimate vehicle weight directly using CV techniques. These include methods that analyze tire deformation \citep{zhang_noncontact_2023, kong_non-contact_2022, feng_application_2021} or bridge deck deformation \citep{ojio_contactless_2016, he_non-contact_2024}, followed by the application of physics-based models to infer the associated loads. However, such approaches often face strict requirements for field conditions and are prone to instability in real-world deployments, limiting their practical applicability. Besides, such approaches are sensitive to shooting distance, lighting conditions, and vehicle speed, which makes their deployment on highways challenging \citep{zhang_factors_2024}.

In general, computer vision and deep learning-based vehicle detection systems can achieve high accuracy—typically exceeding 90\% under ideal lighting conditions \citep{premaratne_comprehensive_2023}. However, vision-based approaches face two major categories of challenges: data-related and model-related. On the data side, key issues include the large volume of video data, which imposes significant demands on data transmission and storage; the quality, diversity, and potential bias of training datasets; and, most critically, privacy concerns. Surveillance systems frequently capture sensitive information, such as human faces and vehicle license plates. While edge-based anonymization techniques are often employed to mitigate privacy risks, achieving complete anonymity remains difficult in practice \citep{azfar_deep_2024}. From a modeling perspective, vision-based systems are sensitive to environmental conditions. Occlusion by other objects, as well as reduced visibility during nighttime or adverse weather (e.g., fog, heavy rain, snow), hampers their ability to operate reliably in continuous monitoring scenarios. Although thermal imaging using infrared cameras can partially address nighttime visibility issues \citep{yoo_truck_2024}, the intense brightness of vehicle headlights in grayscale images complicates data annotation. This increases the cost and effort required for creating high-quality labeled datasets, ultimately limiting the scalability of such approaches \citep{premaratne_comprehensive_2023}.

\subsection{Non-vision-based traffic monitoring}
While vision-based approaches—relying on visual data and object detection algorithms—dominate the field of traffic monitoring, a range of non-vision-based sensing technologies have also been developed. These methods do not rely on image or video data and instead utilize alternative physical phenomena to detect and characterize traffic events.
For instance, \citet{ishida_design_2019} explore vehicle counting using a roadside stereo microphone array by estimating the time difference of vehicle sound arrivals. However, they also highlight the challenges of applying acoustic detection in noisy environments, which limit its practical deployment. Other acoustic sensing technologies, such as ultrasonic arrays \citep{li_detection_2019} and distributed fiber-optic acoustic sensors \citep{liu_vehicle_2020, fakhruzi_urban_2025}, have also shown promise in traffic monitoring applications.
Magnetometer-based solutions are another widely studied category due to their compact size and cost-effectiveness. These systems detect disturbances in the magnetic field caused by moving vehicles, enabling vehicle counting and classification. Several studies have reported accuracies exceeding 80\% with single or multiple magnetometers installed roadside or embedded in the pavement \citep{wang_roadside_2018, dong_improved_2018}. In more complex settings involving multi-lane roads, \citet{sun_smart_2025} demonstrate that integrating data from a network of magnetic sensors using machine learning techniques—such as convolutional neural networks (CNNs) and transformers—can further improve accuracy, reducing traffic volume detection errors to approximately 1.6\%.

Sensors commonly used for structural condition assessment—such as accelerometers and strain gauges—have also been widely adopted in traffic monitoring applications. A classic unsupervised method involves estimating vehicle speed and wheelbase by analyzing the number of signal peaks generated by each vehicle and the time intervals between them \citep{ye_monitoring_2017, ye_collecting_2020, burrello_enhancing_2020}. This approach relies on a simple yet effective principle: passing vehicles induce strain responses and vibrations in the bridge structure.
The primary advantage of this method lies in its ease of deployment and lack of requirement for labeled data. However, several limitations constrain its performance. On heavily trafficked bridges, responses from closely spaced vehicles may overlap, complicating signal interpretation. Moreover, the strain signals generated by smaller vehicles can be obscured when traveling alongside heavy trucks. Additionally, this method typically requires placing sensors near the impact zone—such as beneath the bridge deck—which may not align with optimal sensor placement for SHM objectives. Collectively, these factors limit the scalability and robustness of unsupervised approaches in complex, real-world conditions.

Building on the limitations of unsupervised approaches, recent studies have increasingly turned to machine learning techniques for traffic monitoring using SHM sensor data. These methods offer improved accuracy and robustness by learning complex patterns from labeled datasets. For instance, \citet{yoshida_traffic_2021} apply linear discriminant analysis, a supervised learning technique, to detect the presence of passing vehicles based on vibration signals. Similarly, \citet{ye_collecting_2020} employ CNNs to classify vehicle types from sensor readings.
The integration of CV and WIM data has further advanced the field. \citet{hou_cyber-physical_2020} demonstrate a clear correlation between truck weight and mid-span strain responses, establishing an empirical foundation for using strain gauges to estimate vehicle weight. Extending to networked sensing, \citet{li_traffic_2021} utilize graph convolutional networks to fuse data from multiple sensors across a transportation network, outperforming conventional models in long-term traffic volume prediction.
A notable contribution in this direction comes from \citet{burrello_traffic_2022}, who were the first to apply supervised learning models—such as support vector regressors, multi-layer perceptrons, and random forests—to the traffic monitoring problem using SHM sensors. \citet{iacussi_scalable_2026} combined acceleration-based and vision-based data and employed a regression model for vehicle classification and counting, demonstrating the scalability and effectiveness of this approach. However, their study was constrained by a limited dataset, which prevented the exploration of deeper neural architectures that could further enhance predictive performance.

Despite their unique advantages, non-vision-based solutions often face challenges related to deployment, maintenance, and data interpretation. As a result, they are generally perceived to have lower accuracy compared to vision-based approaches, which continue to dominate the field of traffic monitoring. This study aims to bridge this gap by demonstrating that, when combined with supervised learning, non-vision-based methods—specifically those using SHM sensors—can achieve accuracy and operational simplicity comparable to that of state-of-the-art vision-based systems.
%


\section{Methodology}\label{sec3}

\subsection{Problem definition}
\begin{figure}[!htp] 
    \centering
    \includegraphics[width = 0.65\textwidth]{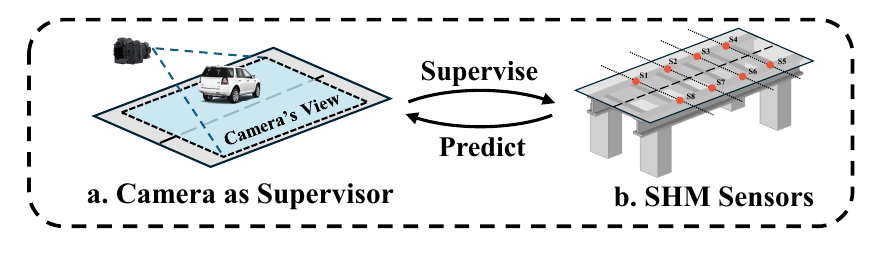}
    \caption{Problem definition}
    \label{fig:require}
\end{figure}

The objective of this study is to develop a traffic monitoring framework that leverages the complementary strengths of vision-based methods and SHM sensors. Specifically, we propose a system that utilizes an existing SHM sensor network installed on the bridge, supplemented by a temporary video recording device, such as a camera or drone, to generate labeled training data (\autoref{fig:require}). Once trained, the system can operate solely on SHM sensor inputs, enabling the continuous monitoring of traffic without the need for permanent visual instrumentation. This approach allows for the repurposing of SHM infrastructure for traffic analysis, reducing the need for additional hardware while maintaining high accuracy and operational efficiency.

Using data from SHM sensors, including acceleration and strain signals, we may formulate traffic volume monitoring as a regression problem. The regression model is designed to take input from the bridge’s sensor network and infer system states, defined as the number of vehicles passing over the bridge within a given time window, categorized by lane and vehicle type. For the purposes of continuous monitoring, we target traffic volume estimation at an hourly resolution.

In our proposed pipeline, outputs from a CV model serve as ground-truth labels for training and validating the regression model. This approach removes the need for manual annotation of sensor signals, which is typically labor-intensive and time-consuming. By comparing the regression model’s predictions against CV-based labels on a held-out test set, we evaluate how well the sensor-based model can replicate the performance of vision-based traffic monitoring. It is important to note that the development and optimization of the CV model itself are beyond the scope of this study.

\subsection{Label generation with computer vision}

To generate labeled training data for the sensor-based regression model, we extract one frame every 20 minutes during the daytime from the surveillance video, forming a mini-dataset for vehicle detection and classification. In each frame, objects are labeled into two categories: light vehicles and heavy vehicles. For this task, we employ YOLO11, a state-of-the-art real-time object detection model \citep{jocher_ultralytics_2024}, which is used to detect and classify vehicles based on their visual appearance.

To enable vehicle tracking across frames, we integrate OC-SORT, a multi-object tracking algorithm that relies purely on motion dynamics \citep{cao_observation-centric_2023}. OC-SORT uses Kalman filter predictions to estimate the future positions of previously detected vehicles and matches these predictions with current detection bounding boxes. This process establishes a consistent identity for each vehicle across multiple frames, allowing us to reconstruct individual vehicle trajectories and ensure accurate counting. With OC-SORT, a trajectory-based consistency check is implemented to minimize the propagation of CV detection errors to the SHM supervision labels. Since single-frame detections may be subject to transient noise, we aggregate the classification results across all frames for a given tracked object ID. The final label is determined via majority voting (i.e., the class predicted most frequently during the object's passage). This temporal consensus mechanism effectively filters out sporadic misclassifications. Furthermore, prior research demonstrates that deep learning models remain robust to substantial amounts of label noise, provided that the clean labels are statistically dominant \citep{rolnick_deep_2018}. By combining the high baseline accuracy of the CV model with this temporal filtering, we ensure that the generated ground truth is sufficiently reliable for supervising the SHM network. Together, YOLO11 and OC-SORT form the core of our CV model, which detects, classifies, and tracks each vehicle in the video stream. In the remainder of this study, this combined system is referred to as the CV model.

The output of the CV model is yielded in the form of bounding boxes in pixel coordinates, which need to be converted into real-world coordinates for traffic modeling purposes. To simplify the pipeline, in this section, we derive a direct, yet approximate, projection without prior camera calibration. Ignoring the edge distortion at the camera’s edges, this process follows a projective transformation, as indicated by \autoref{eq:dlt_1} \citep{faugeras_camera_1992}.

\begin{equation}
k\begin{pmatrix} p_x \\ p_y \\ 1 \end{pmatrix} 
= \begin{pmatrix}
a_{1,1} & a_{1,2} & a_{1,3} & a_{1,4} \\
a_{2,1} & a_{2,2} & a_{2,3} & a_{2,4} \\
a_{3,1} & a_{3,2} & a_{3,3} & a_{3,4}
\end{pmatrix}
\times \begin{pmatrix} X_w \\ Y_w \\ Z_w \\ 1 \end{pmatrix}
\label{eq:dlt_1}
\end{equation}

where $k$ is the third homogeneous coordinate, ($p_x, p_y$) is the pixel coordinate of a point, ($X_w, Y_W, Z_w$) is the world coordinate of a point, and $a_{i,j}$ denotes elements of the transformation matrix, usually determined by the camera's intrinsic and extrinsic parameters.



\begin{figure}[!htp] 
    \centering
    \includegraphics[width = 0.4\textwidth]{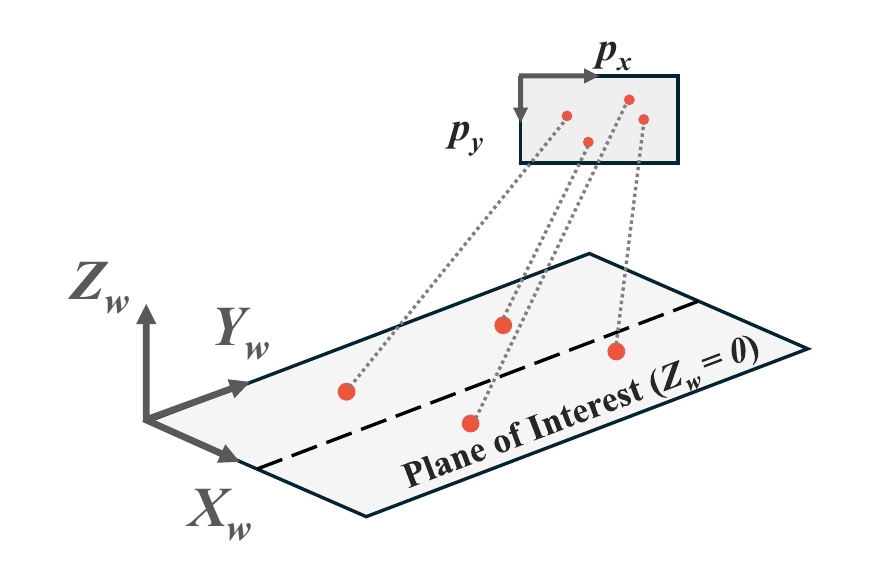}
    \caption{At least four pairs of points on the bridge ($X_w, Y_w$) and corresponding pixels in the video frame ($p_x, p_y$) are used to solve the approximate projection between the plane of interest and video frames. The region of interest within the bridge is assumed to be flat.}
    \label{fig:camera_projection}
\end{figure}

In general, it is not possible to directly back-project individual pixels into real-world coordinates, as each pixel represents all points along a line of sight extending from the camera lens into the scene, irrespective of depth. However, for the specific case of the bridge deck in this study, the road surface is sufficiently flat to be approximated as a planar surface. By defining this plane as $Z_w=0$, the projection problem simplifies to a homography between the image plane and the road plane. This results in a one-to-one mapping between image pixels and real-world coordinates on the defined plane, as shown in \autoref{eq:dlt_2} and illustrated in \autoref{fig:camera_projection}.

\begin{equation}
k\begin{pmatrix} p_x \\ p_y \\ 1 \end{pmatrix} 
= \begin{pmatrix}
a_{1,1} & a_{1,2} & a_{1,4} \\
a_{2,1} & a_{2,2} & a_{2,4} \\
a_{3,1} & a_{3,2} & a_{3,4}
\end{pmatrix}
\times \begin{pmatrix} X_w \\ Y_w \\ 1 \end{pmatrix}
\label{eq:dlt_2}
\end{equation}

The projection matrix can be determined, given $n$ independent points, i.e., $n$ pairs of pixel coordinates ($\mathbf{P}_{3 \times n}$) and corresponding real-world coordinates ($\mathbf{X}_{3 \times n}$). In this case, the transformation can be written as in \autoref{eq:dlt_3}, where $\mathbf{A}_{3 \times 3}$ is the projection matrix and $\mathbf{K}_{n \times n}$ is a diagonal matrix composed of $k$ for $n$ different points. For $n > 4$, the system becomes overdetermined and a least-squares fitting is used to derive the projection matrix, minimizing the squares of the residuals $\mathbf{PK} - \mathbf{AX}$.

\begin{equation}
\mathbf{P}_{3 \times n}\mathbf{K}_{n \times n} = \mathbf{A}_{3 \times 3} \mathbf{X}_{3 \times n}
\label{eq:dlt_3}
\end{equation}

With projection matrix $\mathbf{A}_{3 \times 3}$, the bottom midpoint of the bounding box of vehicles as the reference point is projected to the real world to represent the vehicle's real-time position in the plane of interest. The transformation error occurring in this process mainly depends on the distance between the reference point and the camera \citet{ge_benchmark_2024}. The wide-angle distortion of the camera and the curvature of the bridge surface further amplify the transformation error. The midpoint assumption may introduce bias when determining the precise lateral position of the vehicle. However, these approximations are deemed acceptable for the presented study, as our focus lies on the number of vehicles over a period of time.

The average speed of each vehicle can be obtained by leveraging the projected position and timestamp. Frames with missing detection are interpolated to ensure the continuity of the motion trajectory. To ensure continuity in the motion trajectories, frames with missing detections are interpolated. Additionally, the timestamps corresponding to each vehicle’s entry ($y=0$) and exit from the plane of interest are estimated based on interpolated speeds (\autoref{fig:camera_projection}). As a result, for each tracked vehicle, the following information is extracted from the video footage: vehicle type (light or heavy), estimated speed, time of entry onto the projection plane, and time of exit.

It is important to note that optimizing the performance of the CV model lies outside the scope of this study, as most CV models already achieve an accuracy exceeding 90\% for basic tasks, such as vehicle detection \citep{premaratne_comprehensive_2023}, while tracking algorithms further compensate for missed detections in consecutive frames. Based on our observations from a 10-minute video in our case study, the accuracy of the CV model reaches 98\%. Therefore, our study aims to demonstrate that even when using uncorrected CV model outputs to train a regression model, the deep learning regression model can still achieve high performance on large datasets, effectively filter out errors, and mitigate uncertainties in the labels of the dataset.

\subsection{Signals and label assignments} \label{sec:preprocess}

Firstly, signals from accelerometers and strain gauges should be pre-processed using standard methods, including filtering and detrending. Then, signals are divided into windows of 5 seconds, and the stride between two pieces of signals is 2.5 seconds to enrich the size of the training set. Subsequently, all signals are normalized with the equation described in \autoref{eq:norm}, where $y$ denotes the signal after preprocessing, $y_{norm}$ is the normalized signal, and $y_{\max}$ represents the global maximum absolute amplitude ($y_{\max, \text{acceletaion}}$ for acceleration and $y_{\max, \text{strain}}$ for strain). It is worth mentioning that \( y_{\text{max}} \) is selected empirically and kept constant across the entire dataset, rather than being dynamically computed based on specific time periods (e.g., hourly or daily maxima). This approach ensures consistent scaling across all samples, including those acquired in real-time, and avoids the inconsistency and convergence issues that may arise from using period-specific maxima. As a result, most normalized values fall within the range of approximately \(-1\) to \(1\), though not all are strictly bounded within this interval.



\begin{equation}
y_{norm} = \frac{y}{y_{max}}
\label{eq:norm}
\end{equation}

\begin{figure}[!htp] 
    \centering
    \includegraphics[width = 0.5\textwidth]{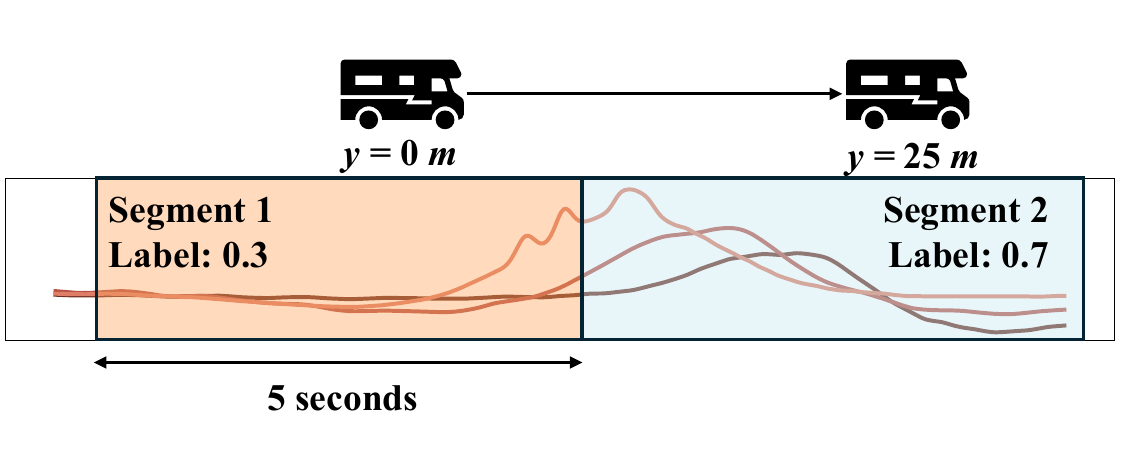}
    \caption{With continuous vehicle counts, each vehicle is proportionally assigned across consecutive signal segments}
    \label{fig:segments}
\end{figure}

For each signal segment, the number of vehicles passing through the sensor network's plane is counted. Importantly, we adopt continuous (non-integer) vehicle counts rather than restricting labels to integer values. For instance, if a vehicle's presence on the plane overlaps with two consecutive time windows, its count is proportionally distributed between the two segments based on the time spent in each window (\autoref{fig:segments}). This labeling strategy ensures that vehicle signals spanning multiple windows are neither duplicated nor omitted, thereby improving labeling consistency and model training quality. The resulting dataset is then randomly divided into training and validation subsets for model development.

\subsection{Feature encoder}
\label{sec:cnn}

The encoder serves as the feature extraction module within the proposed model. We adopt a CNN-based encoder architecture, adapted from the backbone of YOLO11, due to its effective balance between feature representation capacity and computational efficiency. In particular, the inclusion of the C3k2 (Coordinates-To-Features) module enables efficient feature extraction and fusion with minimal latency and resource overhead \citep{jocher_ultralytics_2023, jegham_yolo_2025}. This makes the architecture well-suited for real-time traffic monitoring tasks using SHM sensor signals.

\begin{figure}[!htp] 
    \centering
    \includegraphics[width = 0.75\textwidth]{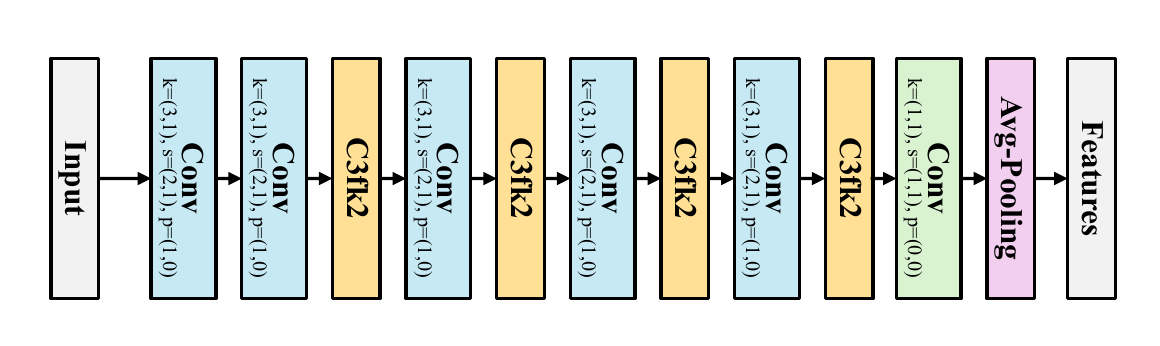}
   \caption{The architecture of the CNN encoder}
   \label{fig:cnn_encoder}
\end{figure}

In the adapted encoder architecture, convolutions are applied exclusively along the time dimension. Although the input signal is two-dimensional—comprising time steps and sensor channels—the CNN performs feature extraction independently for each sensor, without mixing information across the sensor dimension. As a result, the model effectively functions as a 1D CNN applied to each sensor signal in parallel. The architecture of the CNN encoder is illustrated in \autoref{fig:cnn_encoder}. This design choice enhances the model's adaptability and generalizability, as it reduces dependence on the specific configuration or number of sensors, making the pipeline more robust across different deployments.

Another encoder based on feature engineering is also investigated in this study for comparison. Feature engineering is a widely used approach for extracting statistical characteristics from time-series signals. Owing to its relatively low computational requirements compared to deep learning models such as CNNs, this method is particularly well-suited for deployment on edge devices. Running feature extraction locally reduces the need for data transmission to the cloud, thereby lowering communication overhead and enhancing data privacy.
In this study, a set of commonly used time-domain statistical features is extracted from each signal window, including minimum, maximum, mean, standard deviation, kurtosis, root mean square, sum of absolute values, and signal energy. As these features are standard in signal processing, detailed explanations are omitted here. Readers interested in further background are referred to \citet{buckley_feature_2023}.


\subsection{GNN module and multi-task decoder}
\label{sec:gnn}

To effectively capture the spatial dependencies within the SHM sensor network and resolve the signal coupling between adjacent lanes, we model the sensor network as a graph structure. This representation allows the model to leverage the physical topology of the bridge for signal analysis. The sensor network is mathematically modeled as a graph $\mathcal{G} = (\mathcal{V}, \mathcal{E})$. The set of nodes $\mathcal{V} = \{v_1, \dots, v_N\}$ represents the sensors deployed on the structure; in this study, $N$ denotes the number of accelerometers or strain gauges distributed across the bridge cross-section. The set of edges $\mathcal{E}$ describes the connectivity between sensors. Edges are established based on engineering domain knowledge, linking nodes $(v_i, v_j)$ if they are physically connected by structural elements (e.g., beams, columns, or slabs). This topology encodes the expected structural correlations in the captured signals. Each node $v_i$ is initialized with a feature vector $\mathbf{h}_i \in \mathbb{R}^{F_{in}}$, where $F_{in}$ denotes the dimensionality of the feature vector, which is determined by the specific encoder architecture employed (i.e., the CNN-based encoder or the feature engineering encoder). These features are extracted from the raw 5-second signal segments by the encoder described in \autoref{sec:cnn}.

Graph attention networks v2 (GATv2) are applied to perform message passing between sensor nodes \citep{brody_how_2022}; this essentially propagates structural response patterns across the sensor network in our scenario. This mechanism is critical for distinguishing lane-specific loads from global structural vibrations. Unlike static graph convolution that assume fixed edge weights, GATv2 computes dynamic attention coefficients to weigh the importance of neighboring nodes. Specifically, the attention mechanism enables the model to selectively focus on nodes directly excited by a vehicle load while suppressing signals from cross-coupled nodes (e.g., sensors located under an adjacent lane). Following message passing, each node obtains an updated feature vector $\mathbf{h}'_i \in \mathbb{R}^{F_{out}}$ (with $F_{out} = 1024$ in our implementation), which encodes both local signal characteristics and the resolved spatial contextual information. We refer readers to \citet{brody_how_2022} for the comprehensive mathematical derivation.

To translate the node-level features into a global prediction for the entire bridge segment, a global readout function is applied. We employ average pooling to aggregate all node features into a single graph-level representation $\mathbf{h}_{\mathcal{G}}$:

\begin{equation}
    \mathbf{h}_{\mathcal{G}} = \frac{1}{N} \sum_{i=1}^{N} \mathbf{h}'_i
\end{equation}

Finally, to enforce strict lane separation, $\mathbf{h}_{\mathcal{G}}$ is passed through a multi-task decoder. Rather than aggregating all traffic into a single scalar, the decoder branches into separate fully connected layers to regress the final output $\hat{\mathbf{y}}$, representing the specific vehicle count and classification for each lane-category combination. This architectural constraint forces the latent representations to disentangle mixed structural signals into distinct lane assignments.

\subsection{Loss function}
\label{sec:loss}

Before formulating the loss function, it is necessary to define the regression target. Let the ground truth for a given signal window be denoted as a vector $\mathbf{y} \in \mathbb{R}^{K}$, where $K$ represents the number of vehicle categories. In this study, we set $K=4$, corresponding to light and heavy vehicles in both the slow and fast lanes. Accordingly, the model prediction is denoted as $\hat{\mathbf{y}} \in \mathbb{R}^{K}$. The optimization objective is to minimize the discrepancy between $\mathbf{y}$ and $\hat{\mathbf{y}}$ across the dataset.

In regression tasks, the choice of loss function plays a critical role in guiding model optimization. Two of the most commonly used loss functions are mean absolute error (MAE, also known as L1 loss) and mean squared error (MSE, also known as L2 loss). The selection between them has been widely discussed in the literature \citep{willmott_advantages_2005}. MSE is generally preferred when the underlying error distribution is approximately normal, as it penalizes larger errors more heavily \citep{hodson_root-mean-square_2022, chai_root_2014}. In contrast, MAE is less sensitive to outliers and offers greater robustness in datasets with irregular noise or skewed distributions \citep{hodson_root-mean-square_2022}. They are generally defined as:

\begin{equation}
    \text{MAE} = \frac{1}{B} \sum_{i=1}^{B} |y_i - \hat{y}_i|
\label{eq:mae}
\end{equation}
\begin{equation}
    \text{MSE} = \frac{1}{B} \sum_{i=1}^{B} (y_i - \hat{y}_i)^2
\label{eq:mse}
\end{equation}

where $B$ is the batch size (number of samples), $y_i$ is the true value, and $\hat{y}_i$ is the prediction. We adopt MSE as the foundational metric in our pipeline owing to its advantages in terms of computational efficiency and its smooth gradient near zero, which facilitates convergence.

Another important consideration in the design of the loss function is the imbalance across prediction categories. Since MSE is a scale-sensitive metric, its magnitude can be disproportionately influenced by the frequency of each class in the training data. For example, light vehicles (cars) constitute the majority of the dataset and therefore contribute more significantly to the total MSE. In contrast, heavy vehicles in the left (fast) lane are relatively rare, resulting in a much smaller MSE contribution—even if they are consistently mispredicted as zero.
This imbalance can lead to biased optimization, where the model underperforms on less frequent classes. To address this, we adopt a weighted aggregation of the individual losses across the four vehicle categories. Specifically, we employ an uncertainty-based weighting strategy proposed by \citet{kendall_multi-task_2018}, which dynamically adjusts the contribution of each task (or category) to the total loss. By modeling the output uncertainty using a Gaussian assumption, the optimization objective can be formulated as maximizing the log-likelihood of the predictions—or, equivalently, minimizing the corresponding negative log-likelihood—as shown in \autoref{eq:uncertainty_loss}:

\begin{equation}
\mathcal{L} = \sum_{k=1}^{K} \left( \frac{1}{2\sigma_k^2} (y_i - \hat{y}_i)^2 + \log \sigma_k^2 \right)
\label{eq:uncertainty_loss}
\end{equation}

where $K$ denotes the number of vehicle categories and $\sigma_k$ is a trainable parameter and represents the model’s observation noise for category $k$. The term $\log \sigma_k$ serves as a regularizer. During training, the model learns to increase $\sigma_k$ (down-weighting the loss) for tasks with high uncertainty and decrease it for stable tasks. This mechanism eliminates manual weight tuning and ensures that underrepresented classes—such as heavy vehicles in the fast lane—are effectively learned despite their smaller contribution to the raw MSE. For readers interested in mathematical derivation, please refer to the research by \citet{kendall_multi-task_2018}.

\subsection{Evaluation}
\label{sec:evaluation}

While MSE is advantageous during model training due to its sensitivity to large errors and favorable convergence properties, it is less interpretable from a physical perspective. In contrast, MAE offers a more intuitive interpretation, especially in the context of traffic monitoring, where it directly reflects the average deviation between predicted and actual vehicle counts. Following common practice in related work \citep{burrello_traffic_2022, liu_vehicle_2020}, we therefore adopt MAE as the primary metric for evaluating model performance, as it clearly indicates how many vehicles are miscounted over a given time period.

\begin{figure}[!htp] 
    \centering
    \includegraphics[width = 0.3\textwidth]{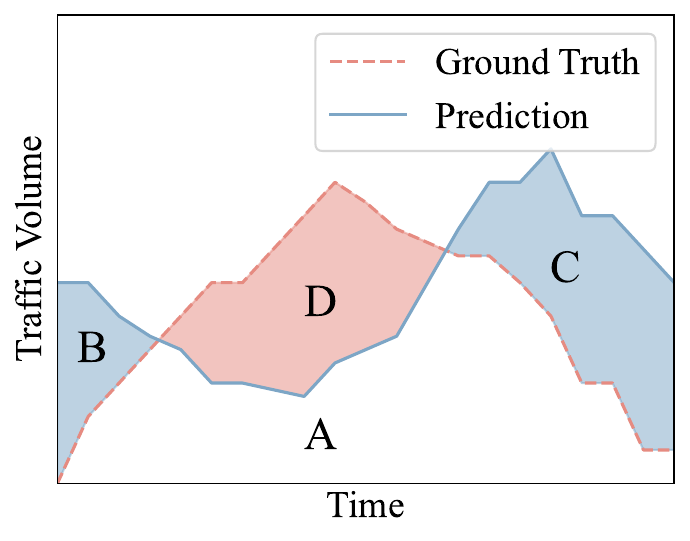}
    \caption{Definition of: True positive: A; False positive: B, C; False negative: D. The larger the overlap area $A$ relative to the total area, the better the alignment between the predicted and true values.}
    \label{fig:accuracy}
\end{figure}

However, as previously noted, while MAE provides an intuitive and interpretable absolute error metric, it remains sensitive to the scale and distribution of the output categories. This can result in disproportionately high error values for majority classes (e.g., light vehicles) and deceptively low errors for minority classes (e.g., heavy trucks), simply due to sample imbalance. To enable fairer cross-category comparisons, it is therefore important to incorporate relative evaluation metrics.
One such metric is the relative MAE (denoted as MAE\%), which normalizes the absolute error by the mean of the corresponding ground truth values. In addition, we recommend the use of a generalized accuracy metric, particularly for long-term time-series data, as introduced in \autoref{eq:accuracy} and illustrated in \autoref{fig:accuracy}. This generalized accuracy is conceptually similar to the intersection-over-union (IoU) measure, and evaluates the degree of overlap between the predicted and actual time-series curves, providing a holistic assessment of the model’s ability to capture traffic dynamics over time.

\begin{equation}
\text{Accuracy} = \frac{TP}{TP+FP+FN} = \frac{A}{A+B+C+D}
\label{eq:accuracy}
\end{equation}

where $TP$ is true positive, $FP$ is false positive and $FN$ is false negative. True negative examples are not applicable or considered zero, as the model predicts the traffic volume, not the absence of vehicles.

With \autoref{fig:accuracy}, MAE in \autoref{eq:mae} simplifies to:

\begin{equation}
\text{MAE} = \frac{B+C+D}{n}
\label{eq:mae_2}
\end{equation}

where $n$ is the number of samples. 





\subsection{Pipeline}

\begin{figure}[!htp] 
    \centering
    \includegraphics[width = 0.6\textwidth]{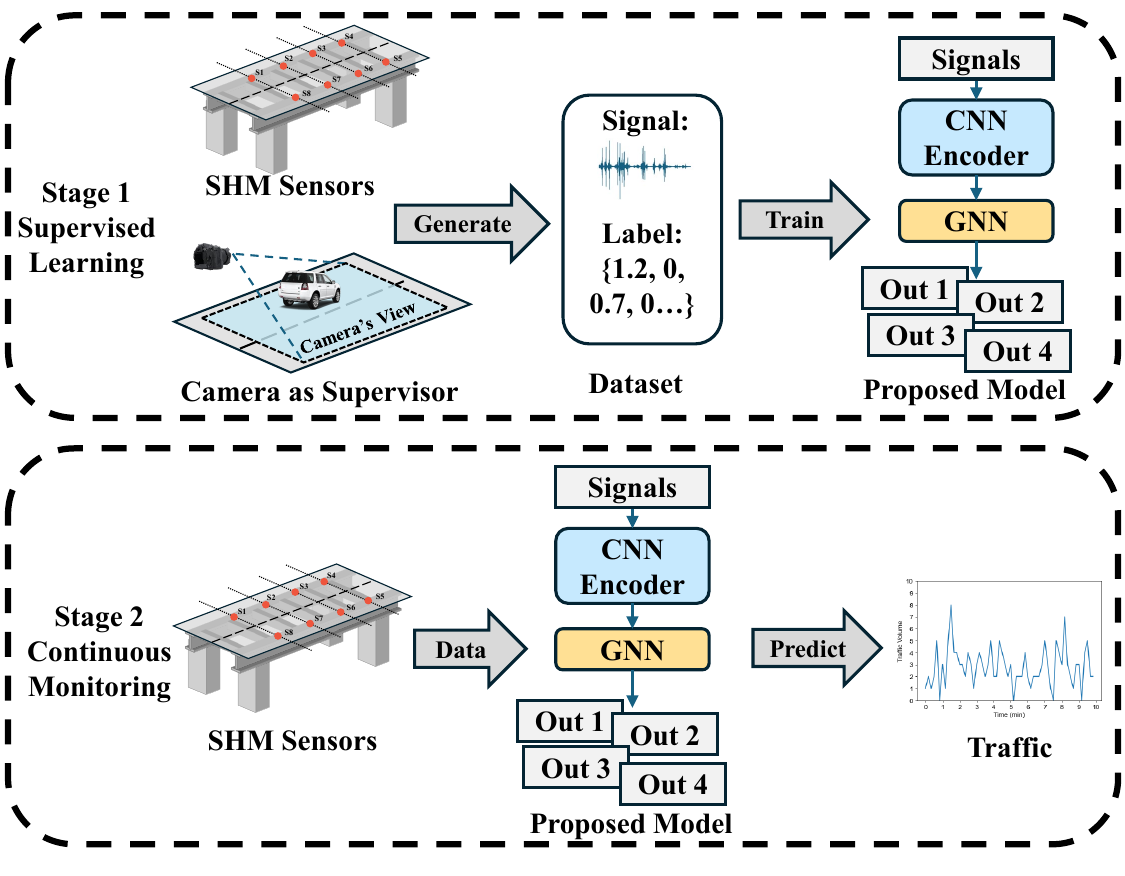}
    \caption{The proposed pipeline}
    \label{fig:pipeline}
\end{figure}

The overall pipeline of the proposed approach is illustrated in \autoref{fig:pipeline}. It consists of two main stages:
\begin{enumerate}
    \item In Stage 1, a temporary video recording device—such as a camera or drone—is deployed on-site for a period of several hours or days to capture traffic footage. These videos are then processed using the CV model, which detects, classifies, and tracks vehicles. The resulting annotations are synchronized with sensor signals (e.g., acceleration and strain) to generate a labeled dataset tailored to the specific bridge.

    \item In Stage 2, once the regression model has been trained on this dataset, the camera is no longer needed. The model can then autonomously estimate traffic volumes using only the signals from the SHM sensors, enabling continuous, camera-free monitoring.
    
\end{enumerate}

\section{Case Study}\label{sec4}

\subsection{Measurement setup}

\begin{figure}[!htp]
  \centering
  \begin{subfigure}{11cm}
    \centering
    \includegraphics[width=0.8\textwidth]{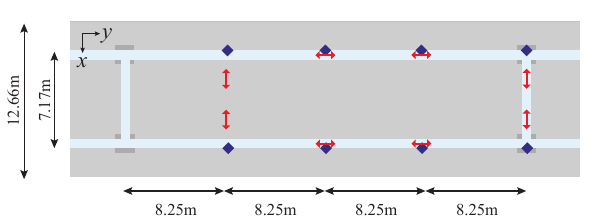}
    \caption{top view}
  \end{subfigure}
  \vspace{1em}
  \begin{subfigure}{11cm}
    \centering
    \includegraphics[width=0.8\textwidth]{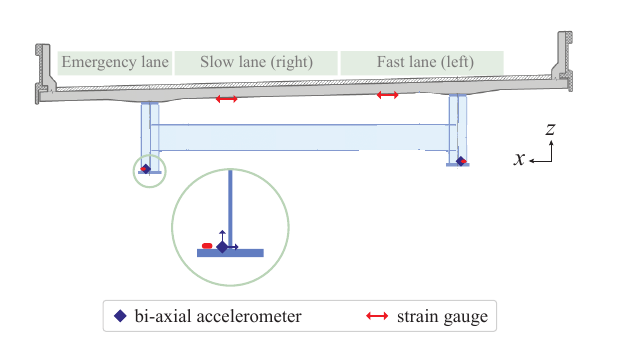}
    \caption{cross section}
  \end{subfigure}
  \caption{Overview of the bridge geometry and sensor network in top view (top) and cross section (bottom).}
  \label{fig:vdc-geometry}
\end{figure}

The data used in this study were collected from a typical Swiss highway bridge. The bridge carries two traffic lanes and one emergency lane, all oriented in the same direction. Structurally, it consists of two main steel girders supporting a post-tensioned reinforced concrete deck slab. Commissioned in 1977, the bridge spans a total length of approximately 370 meters with a typical span length of 32.75 meters. The geometry of the bridge is shown in \autoref{fig:vdc-geometry}. In the daytime, the unidirectional traffic volume averages over 2,000 vehicles per hour. \autoref{sec:statistics} provides further discussion on the bridge traffic.

\begin{figure}[!htp] 
    \centering
    \includegraphics[width = 13cm]{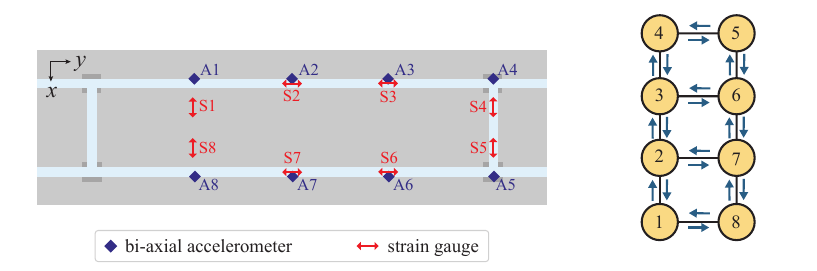}
    \caption{Sensor network layout and its corresponding graph topology for GNN modeling. (Left) The bridge span is instrumented with 8 accelerometers (A1--A8) and 8 strain gauges (S1--S8). Strain gauges S1, S4, S5, and S8 are mounted transversely beneath the concrete slab, while all other strain gauges are mounted longitudinally along the bottom flange of the steel girders. The \(y\)-axis indicates the direction of vehicle travel. (Right) The constructed graph representation where nodes (1–8) denote the physical locations of the accelerometers and strain gauges. The directed edges illustrate the message-passing scheme between neighboring sensors, capturing both longitudinal (vertical arrows) and transverse (horizontal arrows) spatial dependencies.}
    \label{fig:sensor_network}
\end{figure}

A SHM sensor network including eight accelerometers and eight strain gauges is installed below the road plane of the bridge. The installation locations of the sensors and graph representation are shown in \autoref{fig:sensor_network}. The bi-directional accelerometers record data along the vertical and transversal axes at a sampling rate of 250~Hz, while the strain gauges record at 100~Hz. All sensors are hardwired to a central data acquisition unit, which handles both data storage and transmission. For the graph-based representation used in this study, nodes correspond to individual sensors. Edges are defined based on engineering domain knowledge, connecting nodes whose signals exhibit structural correlation. Each node is represented by a feature vector of length 1024, derived from the processed signal segments. Edge features are not utilized in this study.

\begin{figure}[!htp] 
    \centering
    \includegraphics[width = 0.5\textwidth]{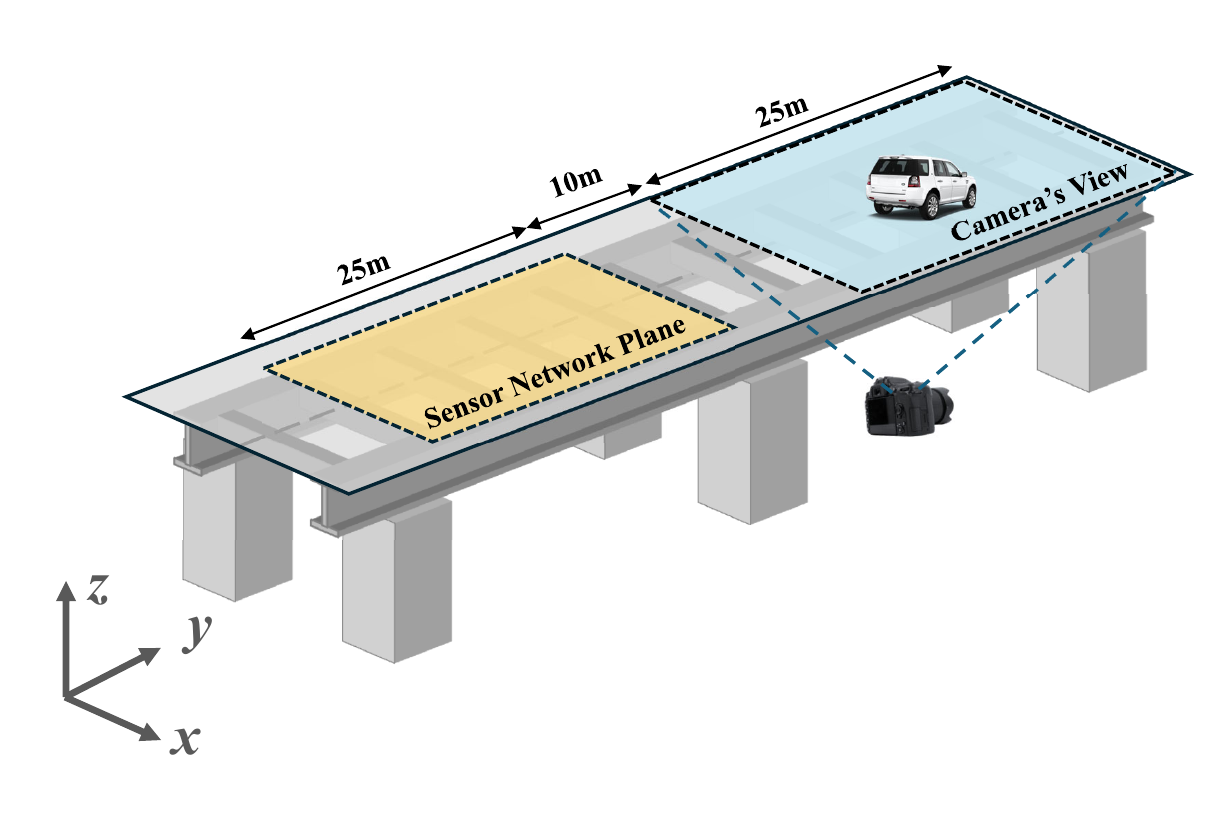}
    \caption{On-site system overview. Accelerometers and strain gauges are installed either beneath the bridge deck or along the steel beams, positioned below the sensor network plane. The surveillance camera is mounted 35 meters downstream of the sensor network, with its field of view aligned in the direction of vehicle travel. The \(y\)-axis corresponds to the direction of traffic flow.}
    \label{fig:onsite_system}
\end{figure}

A temporary surveillance camera is also installed next to the sensor network plane to capture real-time traffic information (\autoref{fig:onsite_system}). The camera is positioned at the downstream end of the SHM sensor network, with its orientation aligned with the direction of vehicle travel. This configuration was dictated by physical constraints related to the location of the data acquisition unit and cable length, as well as by data privacy considerations, as it avoids recording drivers’ faces in oncoming traffic. While this setup induces a temporal offset between the sensor signals and video footage, potentially leading to labeling misalignment owing to vehicle lane changes, such stochastic noise remains well within the robust learning capabilities of the proposed framework \citep{rolnick_deep_2018}. Once the data labeling is completed, the camera will be removed.

\subsection{Label processing}

\begin{figure}[!htp] 
    \centering
    \includegraphics[width = 0.4\textwidth]{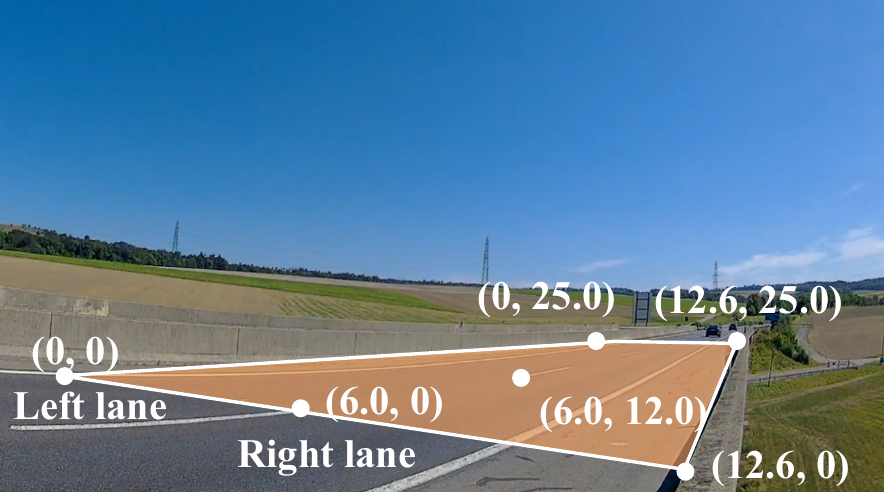}
    \caption{Control points and coordinate system definition on the interested plane}
    \label{fig:control_pts}
\end{figure}

In this study, we set the point next to the left lane as the origin of our system. All other control points, for example, the width of the lane and the spacing of the dashed lines in the middle of the road, are also known with engineering documents and regulations (\autoref{fig:control_pts}). We chose a length of 25 meters for the plane of interest, because the sensor network is also arranged over a 25-meter plane, only shifted by 35 meters (\autoref{fig:onsite_system}).

Due to the fact that the sensor network plane is 35 meters ahead of the camera's view, vehicle events from the downstream camera and upstream sensor signals need to be synchronized. Since the camera is positioned 35 meters downstream from the sensor network, a dynamic time-alignment procedure is implemented. For each vehicle tracked by the CV model, its estimated speed is used to compute the exact time gap required for the vehicle to travel from the sensor network to the camera’s field of view, ensuring that the structural response is accurately mapped to the visual label. This process mitigates label noise but remains sensitive to significant speed fluctuations or lane changes occurring within the 35-meter gap.

It is important to note that the CV model occasionally classifies two-axle trucks as heavy vehicles, even though their weight and the structural response they induce are more comparable to those of light vehicles. To address this discrepancy in our case study, a post-processing step is applied: only vehicles with three or more axles are labeled as heavy vehicles, ensuring better alignment between visual classification and the actual structural impact captured by the SHM sensors.

\subsection{Dataset generation}

Using the installed on-site system, we collected continuous data over a one-week period in August 2024. Video recordings captured during nighttime hours—when vehicles are difficult to identify—were excluded from the dataset. As a result, the final dataset comprises 70 hours of synchronized sensor and camera data, averaging 10 hours per day. Data from five days were used for training and validation, while the remaining two days were reserved as an unseen test set to ensure unbiased performance evaluation.

To ensure temporal alignment and signal consistency, the raw data from the eight accelerometers (originally sampled at 250~Hz) are downsampled to a target frequency of $f_s = 100$~Hz using a polyphase filtering method. These signals are linearly detrended and normalized by a scaling factor $y_{\max, \text{acceleration}}$, which corresponds to the global maximum absolute amplitude of acceleration. Simultaneously, the raw strain gauge signals undergo a multi-stage preprocessing pipeline to eliminate zero drift and high-frequency noise. First, a linear trend is removed. Subsequently, a symmetric exponentially weighted moving average is applied to estimate and subtract the dynamic baseline drift. Let $x[t]$ denote the input strain signal at time step $t$. The forward and backward smoothing sequences, denoted as $\mu_f[t]$ and $\mu_b[t]$ respectively, are calculated recursively with a smoothing factor $\alpha = 0.1$:

\begin{equation}
    \mu_f[t] = \alpha x[t] + (1 - \alpha) \mu_f[t-1], \quad \mu_b[t] = \alpha x[t] + (1 - \alpha) \mu_b[t+1]
\end{equation}

The baseline drift $d[t]$ is estimated as the average of these bidirectional passes to strictly preserve phase information: $d[t] = \frac{1}{2} (\mu_f[t] + \mu_b[t])$. The drift-corrected signal, $\hat{x}[t] = x[t] - d[t]$, is then processed through a second-order Butterworth low-pass filter with a cutoff frequency of $f_c = 2.5$~Hz to extract quasi-static structural responses. Finally, the processed strain signals are normalized by $y_{\max, \text{strain}}$. Both sensor modalities are concatenated and segmented into synchronized windows, yielding a final tensor shape of $(8, 500)$.

Using the data from the first five days, a total of 69{,}849 signal segments are generated, following the preprocessing steps described in \autoref{sec:preprocess}. The resulting dataset is randomly divided into training and validation sets with a 80\%/20\% split. To improve model robustness and prevent overfitting, Gaussian noise \( \epsilon \sim \mathcal{N}(0, 0.04) \) is added to the training set for data augmentation. For unbiased evaluation, the remaining two days of data are used exclusively for testing. These test segments are generated using the same window size but without overlap. A summary of dataset specifications and experimental setup is shown in \autoref{tab:dataset_specs}.

\begin{table}[htp!]
  \centering
  \caption{Summary of dataset specifications and experimental setup.}
  \label{tab:dataset_specs}
  \renewcommand{\arraystretch}{1.2}
  \begin{tabular}{ll}
    \toprule
    \textbf{Parameter} & \textbf{Specification} \\
    \midrule
    \multicolumn{2}{l}{\textit{Sensor Configuration}} \\
    \quad Sensor Nodes & 8 Accelerometers, 8 Strain Gauges \\
    \quad Raw Sampling Rate & 250 Hz (Accelerometers), 100 Hz (Strain Gauges) \\
    \quad Target Sampling Rate & 100 Hz (Resampled \& Synchronized) \\
    \midrule
    \multicolumn{2}{l}{\textit{Preprocessing \& Windowing}} \\
    \quad Window Duration & 5.0 seconds (500 time steps) \\
    \quad Training Stride & 2.5 seconds (50\% Overlap for Augmentation) \\
    \quad Testing Stride & 5.0 seconds (No Overlap) \\
    \quad Input Dimension & $8 \times 500$ (Nodes $\times$ Time steps) \\
    \midrule
    \multicolumn{2}{l}{\textit{Data Partitioning}} \\
    \quad Partition Strategy & First 5 Days vs. Last 2 Days \\
    \quad Samples (Train \& Val) & 69{,}849 segments (First 5 Days) \\
    \quad Train/Val Ratio & 80\% / 20\% (Randomized) \\
    \quad Test Set & Last 2 Days (Excluded from split) \\
    \bottomrule
  \end{tabular}
\end{table}

\subsection{Dataset statistics}
\label{sec:statistics}

The vehicle count results for daytime periods over the course of the week, as estimated by the CV model, are presented in \autoref{tab:tlm_cv}. The data reveal that heavy vehicles (e.g., trucks) predominantly travel in the right lane, while light vehicles are more evenly distributed across lanes. These four traffic categories—defined by vehicle type (light or heavy) and lane (left or right)—serve as the regression targets in this study.

\begin{table}[!htp] 
\begin{center}
    \begin{tabular}{c|cc}
    \hline
     & Left Lane  & Right Lane \\ 
    \hline
    Light vehicles  & 55572  & 68498 \\ 
    Heavy vehicles  & 52     & 4784 \\ 
    \hline
    \end{tabular}
    \caption{Traffic volume statistics over one week, recorded between 10:00 AM and 8:00 PM, as estimated by the CV model. The left lane, designated as the overtaking lane, is located farther from the camera, while the right lane is closer to the camera.}
\label{tab:tlm_cv}
\end{center}
\end{table}

\begin{figure}[!htp] 
    \centering
    \includegraphics[width = 0.4\textwidth]{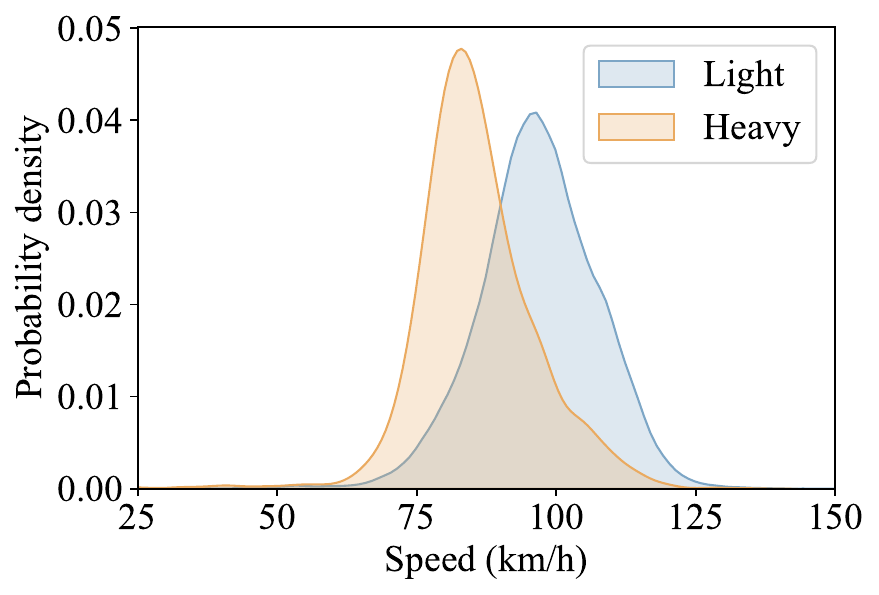}
    \caption{The distribution of speeds of light and heavy vehicles on the bridge}
    \label{fig:speed}
\end{figure}

The speeds of light and heavy vehicles follow normal distributions, denoted as $\mathcal{N}(96.5, 11.7)$ and $\mathcal{N}(85.0, 13.1)$ (unit: km/h), respectively, as demonstrated in \autoref{fig:speed}. Given a span length of $30\,\text{m}$, the number of captured frames per vehicle is calculated to be 28 on average (at $96\,\text{km/h}$) and remains at least 20 frames even at higher speeds ($131\,\text{km/h}$). Combined with the high sampling rates of the SHM sensors, this ensures that all vehicles appearing within the field of view are reliably captured.

\section{Results}

\subsection{Implementation details}

The model is implemented using PyTorch~2.3.1 with CUDA~12.1 and the Deep Graph Library (DGL)~2.4.0, and trained on a single NVIDIA GeForce RTX~4090 GPU. The dataset is randomly partitioned into training and validation sets with an 80:20 ratio. We adopt the AdamW optimizer for model training, utilizing gradient clipping with a maximum norm of 5.0 to stabilize the gradient updates. A cosine annealing learning rate schedule with warm restarts is employed, featuring a 30-epoch warm-up phase reaching a peak of 0.005, followed by a decay cycle over 100 epochs to a minimum of $1 \times 10^{-7}$. Training is performed with a batch size of 256 over a maximum of 200 epochs. An early stopping criterion is applied to prevent overfitting based on validation performance.

\begin{figure}[!htp] 
    \centering
    \includegraphics[width = 0.4\textwidth]{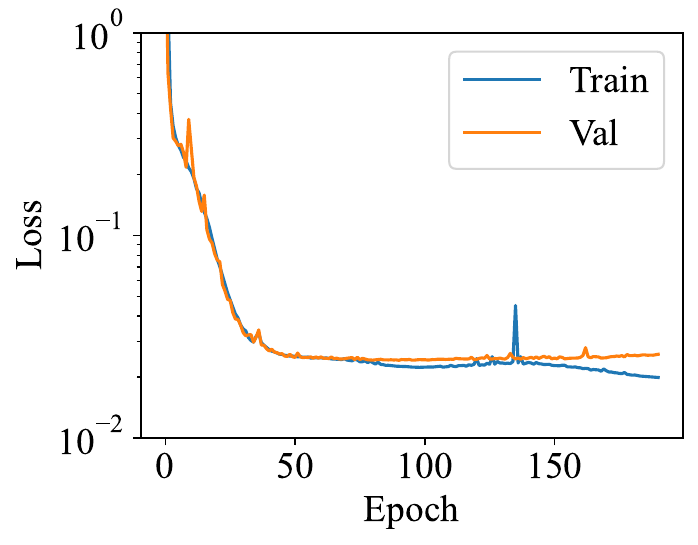}
    \caption{Training and validation loss}
    \label{fig:loss}
\end{figure}

\subsection{Results evaluation}

\begin{figure}[!htp] 
    \centering
    \includegraphics[width = 0.7\textwidth]{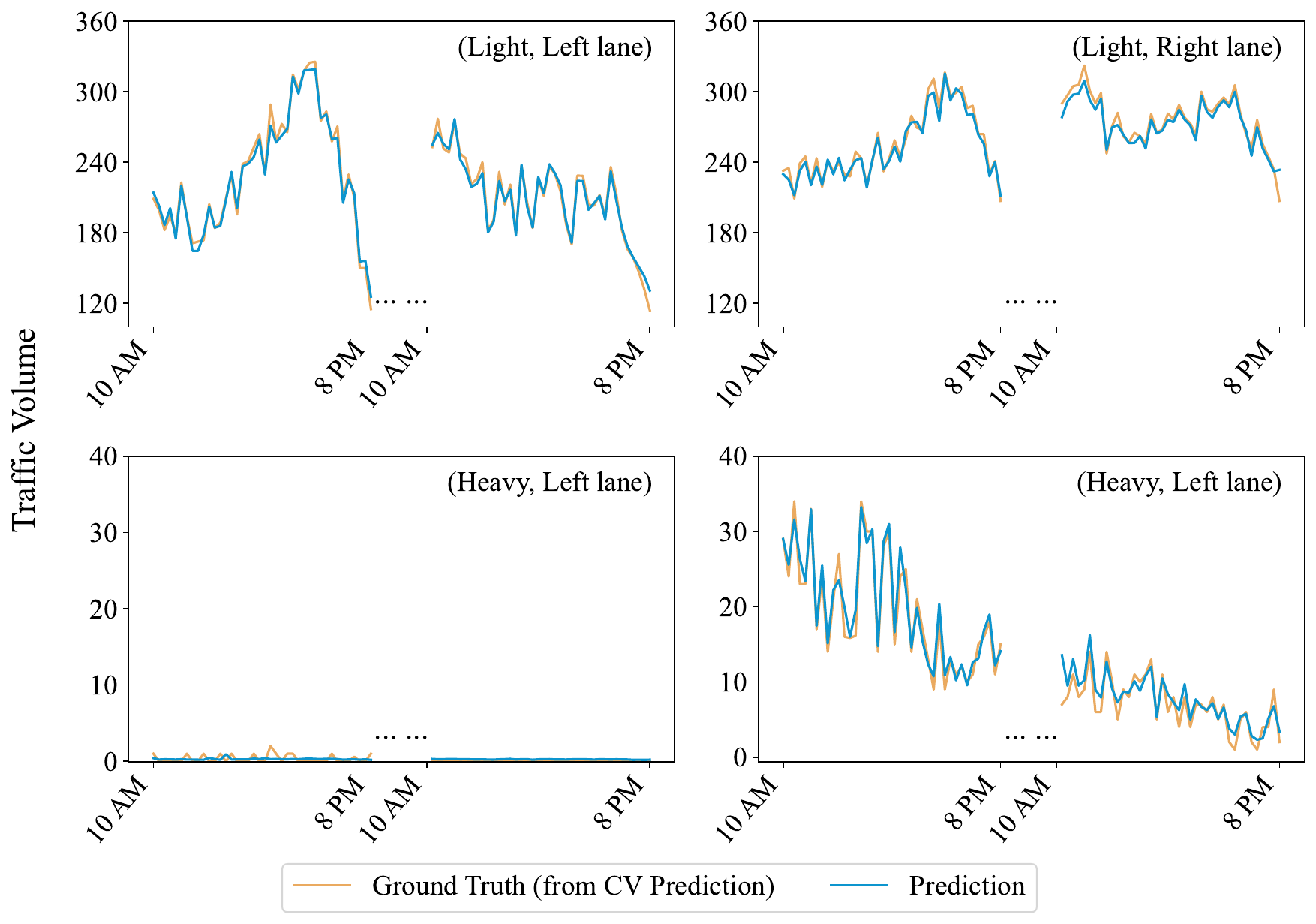}
    \caption{Traffic prediction over two days, with the input signal of two channels (acceleration and strain). The left lane, the overtaking lane, is farther from the camera.}
    \label{fig:dual_hour}
\end{figure}

We processed 20 hours of daytime SHM sensor data collected over two days using the proposed model. The regression predictions for each traffic category are presented in \autoref{fig:dual_hour}.

As a baseline, we use the results obtained from the peak detection method applied to the strain signals, following the approach described in \citet{ye_monitoring_2017, ye_collecting_2020, burrello_enhancing_2020}. The performance metrics presented in \autoref{tab:metrics_r} and \autoref{tab:metrics_l} support the following observations:

\begin{itemize}
    \item The results demonstrate that the model achieves consistently high accuracy using either acceleration or strain as input, validating the effectiveness and flexibility of the proposed pipeline.
    \item Among all categories, light vehicles exhibit the highest prediction accuracy, reaching up to 99\%. Heavy vehicles in the slow lane also achieve strong performance with an accuracy of 94\%. However, the model fails to accurately predict heavy vehicles in the overtaking lane. This can be attributed to the extreme class imbalance—only 40 out of 69{,}849 samples (less than 1\,\textperthousand) belong to this category. As a result, the model incurs minimal penalty for consistently predicting zero, ultimately leading to poor convergence for this class.
\end{itemize}

\begin{table}[!htp] 
\begin{center}
    \begin{tabular}{c|ccc|ccc}
    \hline
    Inputs & \multicolumn{3}{c|}{Light Vehicles} & \multicolumn{3}{c}{Heavy Vehicles} \\ 
    & MAE & MAE\% & Accuracy & MAE & MAE\%  & Accuracy\\
    \hline
    Baseline (Peak Detection) & 139.55 & 13.15\% & 87.13\%  & 24.75 & 46.61\% & 67.94\% \\
    \hline
    Acceleration \& Strain & 9.69 & 0.91\% & \textbf{99.09\%}  & 2.81 & 5.29\% & \textbf{94.83\%} \\
    Acceleration & 15.23 & 1.43\% & 98.58\% & 3.51 & 6.61\% & 93.59\% \\
    Strain & 14.18 & 1.34\% & 98.67\% & 2.92 & 5.50\% & 94.63\% \\
    \hline
    \end{tabular}
    \caption{Evaluation metrics of hourly traffic prediction (right slow lane)}
\label{tab:metrics_r}
\end{center}
\end{table}

\begin{table}[!htp] 
\begin{center}
    \begin{tabular}{c|ccc|ccc}
    \hline
    Inputs & \multicolumn{3}{c|}{Light} & \multicolumn{3}{c}{Heavy} \\ 
    & MAE & MAE\% & Accuracy & MAE & MAE\%  & Accuracy\\
    \hline
    Baseline (Peak Detection) & 106 & 12.08\% & 88.67\% & 0.55 & 84.62\% & 35.29\% \\
    \hline
    Acceleration \& Strain & 18.01 & 2.05\% & 97.98\% & 0.62 &  91.78\% & 36.74\% \\
    Acceleration & 33.07 & 3.77\% & 96.30\% & 0.59 & 86.51\% & 27.99\% \\
    Strain & 10.04 & 1.14\% & \textbf{98.86\%} & 2.17 & 318.87\% & 23.87\% \\
    \hline
    \end{tabular}
    \caption{Evaluation metrics of hourly traffic prediction (left overtaking lane)}
\label{tab:metrics_l}
\end{center}
\end{table}

\subsection{Ablation study}

To evaluate the contribution of each component in the proposed architecture, we conduct an ablation study by comparing several model variants:

\begin{itemize}
    \item \textbf{CNN encoder + GNN decoder}: the full proposed model, using a convolutional encoder followed by a graph neural network decoder.    
    \item \textbf{CNN encoder only}: the GNN module is removed, and a fully connected layer is applied directly after the CNN encoder.    
    \item \textbf{Feature Engineering (FE) encoder + GNN decoder}: precomputed statistical features from the signals are used as input node features for the GNN. Details of the extracted features are provided in \autoref{sec:cnn}.   
    \item \textbf{FE encoder + Multilayer Perceptron (MLP) decoder}: statistical features are fed directly into a standard MLP for regression.
\end{itemize}

For clarity, we report results only for light vehicles and heavy vehicles in the slow lane, excluding predictions for heavy vehicles in the overtaking lane due to the extreme class imbalance. The performance of each model variant is summarized in \autoref{tab:ablation}.

\begin{table}[!htp] 
\begin{center}
    \begin{tabular}{cc|cc}
    \hline
     & Signal Type & Light Vehicles  & Heavy Vehicles \\ 
     \hline
    Baseline & Strain & 88.27\%  & 67.94\% \\ 
    \hline
    CNN+GNN  & Acceleration & 98.72\%  & 93.59\% \\ 
    (Our model)  & Strain & \textbf{99.17\%}  & \textbf{94.63\%} \\ 
    \hline
    CNN       & Acceleration & 98.14\%  & 91.34\% \\ 
              & Strain & 97.84\%  & 93.06\% \\ 
    \hline
    FE + GNN  & Acceleration & 96.70\%  & 93.96\% \\ 
              & Strain & 96.53\%  & 92.00\% \\ 
    \hline
    FE + MLP  & Acceleration & 94.46\%  & 92.72\% \\ 
              & Strain& 97.85\%  & 91.54\% \\ 
    \hline
    \end{tabular}
    \caption{Ablation study: hourly accuracy of different models with acceleration or strain as input}
\label{tab:ablation}
\end{center}
\end{table}

Based on the ablation study results, even the simplest model variant—FE combined with a MLP—achieves over 90\% accuracy in traffic prediction, outperforming the unsupervised baseline. This highlights the general effectiveness of supervised deep learning models for traffic monitoring using SHM signals. Among all tested configurations, the proposed model—featuring a CNN encoder and GNN decoder—achieves the highest performance.
It is important to note that the absolute number of light vehicles is significantly greater than that of heavy vehicles (\autoref{tab:tlm_cv}). Consequently, a small percentage drop in prediction accuracy for light vehicles results in a proportionally larger increase in MAE—approximately 20 additional miscounted vehicles per 1\% reduction in accuracy.


\subsection{Computational Cost}

Real-world SHM deployments often necessitate edge computing solutions, which are typically constrained to CPU-based hardware. To assess the practical feasibility of the proposed method within such systems, we evaluated the inference speed of the proposed model alongside other comparative variants. The experiments were conducted on a server equipped with an AMD EPYC~7763 64-Core Processor and an NVIDIA GeForce RTX~4090 GPU. We calculated the mean inference time for processing a one-hour signal, averaged over the 20-hour test set and across both acceleration and strain channels. The results are presented in \autoref{tab:inference_speed}.

\begin{table}[htp!]
\centering
\caption{Average inference time per one-hour sample on CPU and GPU.}
\label{tab:inference_speed}
\begin{tabular}{lcc}
\hline
\textbf{Model} & \textbf{CPU Time (s)} & \textbf{GPU Time (s)} \\ \hline
Peak Detection (Baseline) & 0.18 & 0.12 \\
\hline
CNN+GNN (Our model) & 8.20 & 2.26 \\
CNN & 4.52 & 0.95 \\
FE + GNN & 1.34 & 1.34 \\
FE + MLP & 0.27 & 0.24 \\ \hline
\end{tabular}
\end{table}

The baseline peak detection method exhibits the lowest latency (0.18s) due to its simplicity. The proposed \textbf{CNN+GNN} model incurs a higher computational cost, with an inference time of 8.20 seconds on the CPU and 2.26 seconds on GPU. This increase is expected due to the comprehensive integration of convolutional layers and graph neural networks for deep feature extraction. The inference time for variants containing the FE encoder module was found to be higher on the GPU than for CNN and GNN architectures. One reason is that the statistical feature extraction in the FE encoder is computationally distinct from convolutional operations. It involves statistical calculations that are CPU-intensive and less parallelizable, thus deriving significantly less benefit from GPU acceleration compared to dense matrix operations in CNNs. Additionally, to simulate a real-time processing scenario, we performed inference on individual signals rather than employing batch processing. Consequently, the model could not leverage the parallelization advantages offered by GPU batch inference. The inference latency is negligible relative to the signal length across all models, demonstrating the method's high feasibility for real-time edge computing in practical SHM systems.

\section{Conclusion}

\subsection{Summary}

To maximize the complementary strengths CV algorithms and SHM sensor networks in traffic monitoring, we proposed an automated pipeline. Our case study demonstrates that the pipeline achieves high performance, strong generalizability, and ease of redeployment using a large-scale dataset. The key conclusions of this study are summarized as follows:

\begin{enumerate}
\item \textit{Automation}. We introduced an automated pipeline that transfers the capabilities of CV-based traffic monitoring to SHM sensor networks. In our case study, CV algorithms effectively generate high-resolution traffic datasets with minimal manual intervention. Deep learning regression models trained on these automatically labeled datasets—with inherent uncertainties—achieve accuracy comparable to vision-based systems. Importantly, the proposed approach offers clear advantages over continuous visual monitoring: the SHM sensor network operates reliably at night, respects privacy constraints, and eliminates the need for permanent cameras. While visual data is used during initial training, it is no longer required during deployment.
    
\item \textit{Sensor network interpretation via CNN and GNN}. Despite imperfect synchronization and label noise from the CV model, the pipeline reliably extracts meaningful features from sensor signals. Among the tested models, GNNs demonstrate superior capability in interpreting coupled sensor signals, particularly within civil infrastructure, where spatial interdependencies are common. The model performs well with both acceleration and strain inputs, confirming its robustness across sensor types. Furthermore, the FE + MLP variant achieves over 90\% accuracy, making it suitable for low-computation environments. A promising future direction involves deploying lightweight models on edge devices to improve both data privacy and transmission efficiency.

\item \textit{Industrial benefits}. The findings confirm the high industrial feasibility of the proposed method. By repurposing existing SHM infrastructure for traffic analysis, the system eliminates the recurring costs associated with manual signal annotation and the intrusive installation of road-level sensors. The transition from a vision-supervised training phase to an autonomous, sensor-only monitoring phase allows for continuous operation under all lighting conditions while maintaining driver privacy, a key requirement for public infrastructure projects. This framework provides a cost-effective solution for bridge managers to monitor traffic loads without the maintenance burden of permanent visual instrumentation.

\end{enumerate}

\subsection{Limitations}
\label{sec:limitations}

The proposed framework is architecturally designed to support cross-structure generalization, utilizing parallel feature extraction and graph-based message passing to adapt to varying structural geometries and sensor configurations. Despite this inherent flexibility, the practical implementation of the vision-supervised pipeline is subject to specific constraints regarding the experimental setup and physical deployment.

First, while we assume the CV model provides sufficiently accurate ground truth for training and evaluation, limitations remain in the camera setup used in our case study. The low camera angle increases the likelihood of occlusion, particularly of vehicles in the left lane. However, given the low proportion of heavy trucks in the monitored traffic flow, such occlusion events remain statistically rare in our dataset. Additionally, the camera is positioned several seconds downstream of the sensor network, preventing detection of lane changes that occur prior to entering the camera's field of view. These factors can introduce label noise and misclassification. Although such uncertainties do not dominate the training data, they may bias evaluation results that rely on vision-based annotations.

On the other hand, a precise synchronization between the temporary camera setup and the SHM network is critically required in the pipeline, as it directly determines the quality of the dataset. This imposes specific constraints on the sensor network configuration and camera placement. First, concerning sensor redundancy and fault tolerance, although this study achieved robust results using four sensors per lane, the minimum number of sensors required for model convergence remains to be fully explored, particularly in scenarios involving sensor faults or failures. Moreover, the requirement for spatial compactness dictates that the sensor network must remain sufficiently compact (e.g., confined to a specific bridge deck section) to preserve the spatial correlation of signals across different sensors, which inherently limits scalability when applied to long-span bridges or spatially extensive sensor networks. Finally, to ensure visual-physical alignment, the camera's field of view should ideally overlap with, or be in close proximity to, the sensor network to ensure spatial alignment between the CV-derived labels and SHM data; excessive distance introduces significant label noise due to traffic dynamics (e.g., lane changes or speed variations), further constraining the system's applicability to large-scale networks.

\subsection{Outlook}

Future work will focus on evaluating the transferability and robustness of the proposed pipeline. While the current results demonstrate strong performance for this specific structural geometry and sensor configuration, future studies will aim to test the model on different bridge typologies, including varying span lengths, materials, or structural dynamics. This will allow for a comprehensive assessment of how the proposed pipeline adapts to different sensor configurations, ultimately solidifying its role as a generalized, privacy-preserving alternative to visual monitoring.

\section*{Data Availability Statement}
Data and demonstrative Python codes that implement the proposed approach are openly available at our public GitHub repository: https://github.com/wuhanshuo/SHMSN-Traffic-Monitoring

\section*{Acknowledgements}
We gratefully acknowledge the Maintenance Planning Division of the Federal Roads Office (FedRO), Switzerland, for granting permission to use the monitoring data in this publication.

\section*{Declaration of Competing Interest}
The authors declare that they have no known competing financial interests or personal relationships that could have appeared to influence the work reported in this paper.

\appendix

\setcounter{table}{0}
\renewcommand{\thetable}{A\arabic{table}}

\section{Model Architecture Implementation Details}
\label{app:implementation}

\textbf{Peak Detection (Baseline):} We employ a peak detection algorithm on the signal to count vehicle passages. For comparison, we use optimal thresholds, which are calibrated by minimizing the MAE between predictions and ground truth. Due to the bridge's asymmetric structure and lane distribution, different thresholds are applied to the sensors in each lane.

\textbf{Feature Engineering (FE):} For FE-based baselines, we extract 8 statistics along the temporal axis for each node: \textit{minimum, maximum, mean, standard deviation, kurtosis, RMS, absolute sum, and energy}.

\textbf{Model Architectures:} Other model architecture implementation details are shown in \autoref{tab:model_comparison_appendix}.

\begin{table}[htp!]
    \centering
    \caption{Model Architecture Implementation Details}
    \label{tab:model_comparison_appendix} 
    \renewcommand{\arraystretch}{1.3}
    \begin{tabular}{p{3.0cm}|p{5.0cm}|p{6.0cm}}
    \hline
    \textbf{Model} & \textbf{Architecture Components} & \textbf{Key Hyperparameters} \\
    \hline
    \textbf{Peak Detection} \newline (Baseline) & Rule-based Logic \newline (\texttt{scipy.signal.find\_peaks}) & \textbf{Min Distance:} 0.1s \newline \textbf{Height Ranges:} \newline Left: Light [0.04, 0.4], Heavy [0.4, 1.0] \newline Right: Light [0.035, 0.1], Heavy [0.1, 1.0] \\
    \hline
    \textbf{CNN + GNN} \newline (Our Model) & YOLO-Backbone (C3k2) \newline + GATv2 + Linear & Input Shape: $8 \times 500$ \newline Filter Sizes: \newline [16, 32, 64, 128, 256, 1280] \newline Kernel Sizes: $(3,1)$ \newline Strides: $(2,1)$ \newline GNN Nodes: 8 \newline GNN Heads: 8, Layers: 2 \newline GNN Latent Dim: $8 \times 128$ \\
    \hline
    \textbf{CNN} & YOLO-Backbone (C3k2) \newline + Linear & Input Shape: $8 \times 500$ \newline Filter Sizes: \newline [16, 32, 64, 128, 256, 1280] \newline Kernel Sizes: $(3,1)$ \newline Strides: $(2,1)$ \newline Final Latent Dim: 1280 \\
    \hline
    \textbf{FE + GNN} & GATv2 + Linear & Input: $8 \text{ nodes} \times 8 \text{ stats}$ \newline Nodes: 8 \newline Heads: 8, Layers: 2 \newline Latent Dim: $128 \times 8$ \\
    \hline
    \textbf{FE + MLP} & 3-Layer MLP & Input: 64 ($8 \text{ sensors} \times 8 \text{ stats}$) \newline Layers: [64, 128, 128, 4] \newline Dropout: 0.2 \\
    \hline
    \end{tabular}
\end{table}


\bibliographystyle{elsarticle-num-names}
\bibliography{references}






\end{document}